# Applications of Explainable artificial intelligence in Earth system science

Feini Huang[1,2], Shijie Jiang[2,5], Lu Li[1], Yongkun Zhang[1], Ye Zhang[1], Ruqing Zhang[3], Qingliang Li[4], Danxi Li[1], Wei Shangguan[1,*], Yongjiu Dai[1]

1Southern Marine Science and Engineering Guangdong Laboratory (Zhuhai), Guangdong Province Key Laboratory for Climate Change and Natural Disaster Studies, School of Atmospheric Sciences, Sun Yat–Sen University, Zhuhai 519082, China.

2Department of Biogeochemical Integration, Max Planck Institute for Biogeochemistry, 07745 Jena, Germany.

3Shanwei Meteorological Service, Shanwei, China

4College of Computer Science and Technology, Changchun Normal University, Changchun 130032, China.

5ELLIS Unit Jena, Jena, Germany

*Correspondence: shgwei@mail.sysu.edu.cn; Tel.: +86-13466654629

**Abstract**

In recent years, artificial intelligence (AI) rapidly accelerated its influence and is expected to promote the development of Earth system science (ESS) if properly harnessed. In application of AI to ESS, a significant hurdle lies in the interpretability conundrum, an inherent problem of black-box nature arising from the complexity of AI algorithms. To address this, explainable AI (XAI) offers a set of powerful tools that make the models more transparent. The purpose of this review is twofold: First, to provide ESS scholars, especially newcomers, with a foundational understanding of XAI, serving as a primer to inspire future research advances; second, to encourage ESS professionals to embrace the benefits of AI, free from preconceived biases due to its lack of interpretability. We begin with elucidating the concept of XAI, along with typical methods. We then delve into a review of XAI applications in the ESS literature, highlighting the important role that XAI has played in facilitating communication with AI model decisions, improving model diagnosis, and uncovering scientific insights. We identify four significant challenges that XAI faces within the ESS, and propose solutions. Furthermore, we provide a comprehensive illustration of multifaceted perspectives. Given the unique challenges in ESS, an interpretable hybrid approach that seamlessly integrates AI with domain-specific knowledge appears to be a promising way to enhance the utility of AI in ESS. A visionary outlook for ESS envisions a harmonious blend where “process-based models govern the known, AI models explore the unknown, and XAI bridges the gap by providing explanations”.

# Contents

## 1. Motivations

Earth system science (ESS) provides a comprehensive perspective on our planet by integrating diverse scientific processes and data from different domains, thereby fostering a profound understanding of Earth as a complex, adaptive system (Schellnhuber 1999; Steffen et al., 2020). With the recent proliferation of data observation and synthesis technologies, such as sensors, remote sensing, and reanalysis, a deluge of Earth system data has become available. Artificial intelligence (AI) is a particularly powerful tool for harnessing this wealth of data and unlocking its maximum potential (Li et al., 2023a). Over the past two decades, AI has flourished in the ESS (Reichstein et al., 2019; Xu and Liang, 2021). It offers advantages in capturing nonlinear patterns, synthesizing multi-source data, reducing computational costs, and complementing traditional numerical models. Traditional machine learning (ML) techniques, such as random forest (RF) and support vector machine (SVM), excel in classification and regression problems and nonlinear system descriptions. On the other hand, deep learning (DL) algorithms such as artificial neural networks (ANNs), recurrent neural networks (RNNs), convolutional neural networks (CNNs) and innovative methods such as reinforcement learning, Transformers, generative adversarial networks (GANs), and diffusion models improve prediction accuracy through their complex architectures.

The effectiveness of AI (specifically traditional ML and DL) in prediction and modeling for ESS has generated considerable interest among geoscientists (McGovern et al., 2019). Nonetheless, the black-box nature of AI models presents a significant barrier to understanding the underlying insights they provide. This black-box nature refers to the complexity of AI models that makes the model difficult for humans to interpret (Reichstein et al., 2019). For example, the multi-layered neural structure of DL obscures transparency and interpretability (Sarker, 2021). Nevertheless, AI can serve as a source of new knowledge for ESS due to its superiority in representing non-linear relationships and systemic complexity. When AI is made explainable, the information hidden in these models has the potential to reveal critical insights that might otherwise be overlooked (Behrens & Viscarra Rossel, 2020; Leist et al., 2022; Krenn et al., 2022). The quest for model explanations in AI has led to the emergence of two main strategies (Shen, 2018; Bergen et al., 2019).

One strategy is to adopt or develop AI architectures that inherently provide clarity to their decision processes, such as decision trees (Tulloch et al., 2018), linear model (Qin et al., 2021) and physics-guided neural networks (Lian et al., 2022). The other strategy focuses on applying post-hoc explainable AI (XAI) techniques to uncover the mechanisms behind the outputs of AI. This review focuses primarily on various methods and applications within the field of XAI. XAI can be a useful tool for overcoming the traditional trade-off between performance and explainability in complex AI systems, thereby fostering trust, supporting decision making, diagnosing and improving models, and enabling scientific discovery across ESS disciplines.

XAI was initiated by the Defense Advanced Research Projects Agency in 2015 with the primary goal of making AI's behavior more understandable to humans by providing explanations (Gunning et al., 2019). Humans often make decisions based on verifiable evidence that bolsters their confidence (Gunning et al., 2019); however, in ESS, AI predictions typically provide outcomes without explaining the underlying processes. This opacity impedes trust in AI and is a major concern for regulators and the public, thereby hindering its adoption in predictive applications (Huntingford et al., 2019; Gevaert, 2022; Cowls et al., 2023; Jobin et al., 2019; Nordgren, 2023).

XAI techniques offer a promising way to address interpretability concerns by making AI models more transparent, credible, and understandable. Specifically, it is expected that XAI could provide a solid foundation for decision making when applying complex DL models to predict extreme events and long-term climate change (Nishant et al., 2020; Prodhan et al., 2022). Moreover, for most AI modelers in the ESS, insights derived from XAI serve as an effective guide for diagnosing and debugging models by augmenting data or model building. Further, from the perspective of theoretical scientists in ESS, XAI is also crucial for scientific understanding and discovery. The essence of science in ESS is the acquisition of knowledge about the world, and the generation of hypotheses hidden in data can be effectively learned by AI and made visible by XAI (Irrgang et al., 2020; Gettelman et al., 2022). While traditional ML models such as decision trees have achieved notable success due to their high explanatory power, complex DL models often surpass them, demonstrating greater potential to uncover underlying processes and knowledge. With XAI, it becomes conceivable to unveil the insights embedded in these models.

Since its rise in 2015, XAI has become a thriving area of interest within the AI community and has found practical applications in many fields. By the end of December 2023, more than 9,000 peer-reviewed articles on XAI and related concepts had been published. Figure 1a illustrates the extensive coverage of XAI research in various fields. In particular, XAI methods have been effectively applied in fields such as computer science (Akula et al., 2022; Mankodiya et al., 2022), medicine (Kim et al., 2022; Lauritsen et al., 2020; Novakovsky et al., 2022; van Hilten et al., 2021), and management and construction (Calvaresi et al., 2021; Kute et al., 2021; Park and Yang, 2022; Weber et al., 2023). Furthermore, XAI has made significant progress in theoretical domains such as chemistry (Zhong et al., 2022; Yano et al., 2022), physics (Li et al., 2022a; Mousavi and Beroza, 2022) and biology (Pilowsky et al., 2022). In ESS and its sub-domains (water, environment, geosciences and remote sensing), the number of publications incorporating XAI was 285, with an increasing trend since 2018 (Figs. 1b and 1c).

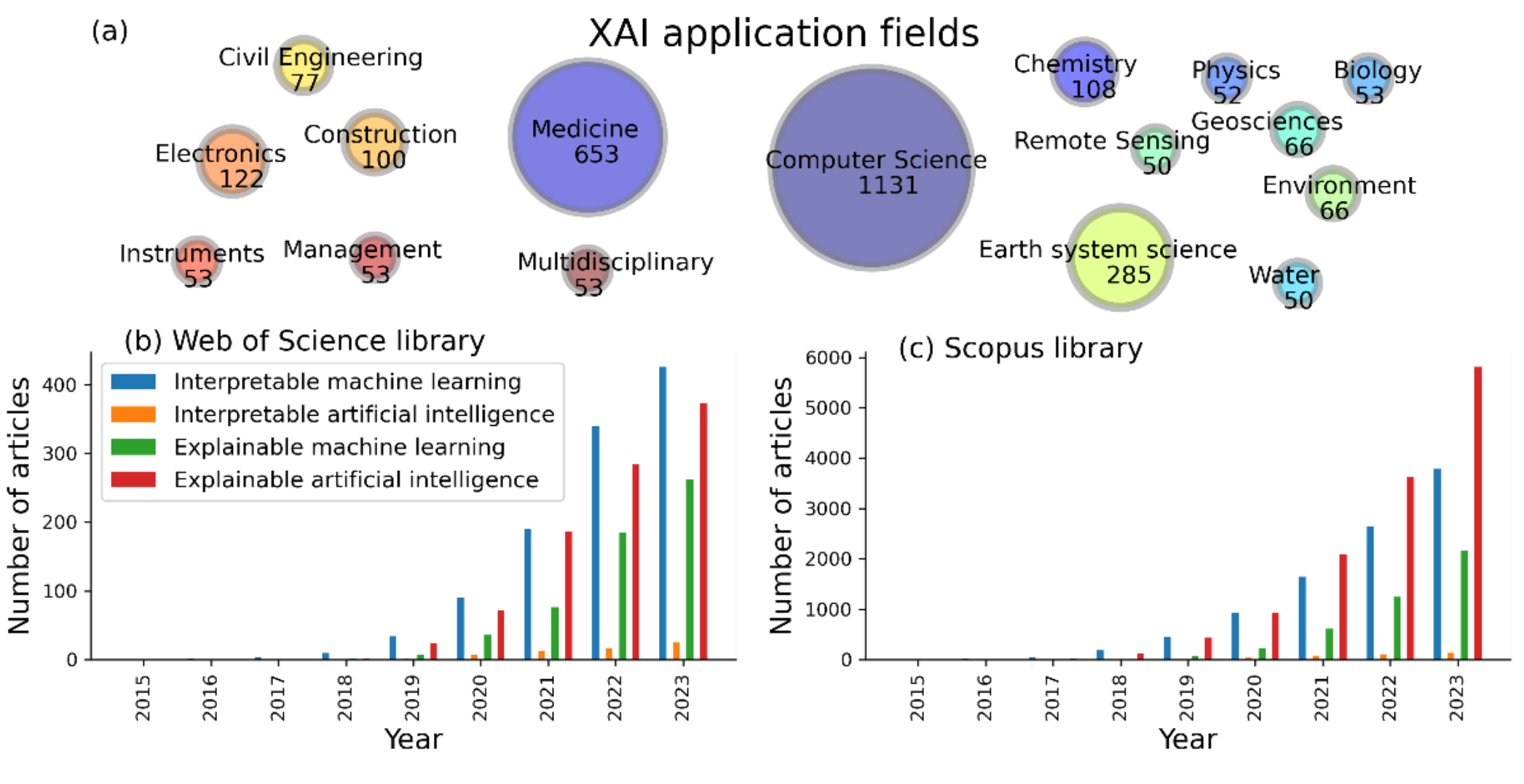

Figure 1. Number of publications related to explainable artificial intelligence (XAI) in various fields. (a) number of publications in XAI application fields; (b) growth trend of XAI publications in Web of Science (2015~2023); (c) in Scopus (2015~2023). Data was retrieved on December 31, 2023, using the search terms listed in the legend for each database query.

Despite the booming interest in XAI in ESS, with more than 285 articles referencing XAI applications in the field, a comprehensive overview of the XAI-ESS interface is still lacking, with the exception of reviews focusing on specific subfields (Başağaoğlu et al., 2022; Gevaert 2022). Although XAI is still in its early stages, many scientists have already recognized its potential to enhance future modeling across ESS components, including meteorology (McGovern et al., 2019), hydrology (Shen, 2018), natural hazards (Dikshit & Pradhan, 2021a, 2021b; Ghaffarian et al., 2023), Earth system monitoring (Tziolas et al., 2021), Earth system modeling (Reichstein et al., 2019), environmental science (Tuia et al., 2021; Zhang et al., 2021a), remote sensing (Camps-Valls et al., 2020; Kakogeorgiou and Karantzalos, 2021), and ecology (Shah et al., 2021). Particularly, Irrgang et al. (2020) have introduced the innovative concept of neural Earth system models (ESM), which aims to integrate AI and ESM in a deep and interpretable way. The European Centre for Medium-Range Weather Forecasts (ECMWF) has planned to incorporate XAI into future modeling efforts with the dual purpose of improving performance and discovering novel natural phenomena that can further inform process-based (PB) models (Schneider et al., 2022). By 2023, XAI has continued its growth trajectory within the ESS, serving the three core goals of communicating model decisions, improving models, and providing scientific insights. Numerous researchers have approached XAI from diverse perspectives. For example, McGovern et al. (2019) introduced XAI to the field of atmospheric science, demonstrating successful applications such as precipitation classification and hail prediction. From a regulatory perspective, Gevaert (2022) highlighted the different explanation requirements of various audiences in Earth observation and proposed a standardized flowchart to address these needs. Başağaoğlu et al. (2022) summarized the use of XAI in different hydroclimatic domains and presented an automated XAI framework. Overall, XAI has the potential to improve ESS by providing interpretable and trustworthy AI models that can drive informed decision-making and scientific discovery.

However, the integration of XAI into ESS faces several challenges. One major obstacle is that many ESS researchers are unfamiliar with XAI methods and are therefore reluctant to adopt them due to their early stage of development. The rapid evolution of available XAI methods, which include a lot of jargon familiar to the ML community, often leaves geoscientists uncertain about the appropriate method for their specific research questions due to the interdisciplinary knowledge gap. This hesitation is exacerbated by the rapidly advancing but early stage of XAI development, where technical issues abound and theoretical foundations are still being established.

In summary, despite the booming application of XAI in ESS, technical challenges and compatibility issues between XAI and ESS models pose significant problems. Technical challenges such as faithfulness, robustness, and stability issues of XAI methods may hinder the application. Also, the mismatch of interpretative forms (e.g., feature importance and latent representation) and prior physical knowledge in ESS would cause the compatibility problems. The high expectations surrounding XAI might bring the unpredictable risk of its potential misuse in ESS.

In this review, our primary goal is to elucidate XAI methodologies for the broader ESS community by providing a comprehensive overview of the current state of XAI applications in ESS. We aim not only to highlight their remarkable achievements in advancing ESS development, but also to explore the challenges and opportunities that lie ahead. Specifically, this paper is organized as follows:

(1) Section 2 provides a brief technical overview of classical and common XAI methods, along with a comparative analysis of their strengths and weaknesses.

(2) Section 3 presents a comprehensive summary of the diverse applications of XAI to different goals in the ESS, detailing how XAI helps to communicate model decisions, diagnose and improve models, and uncover scientific insights.

## 2. XAI Basics

### 2.1 Definition, Approaches and Interpretative Forms

The field of XAI derives from the broader concept of "interpretability" or "explainability". To avoid confusion for the ESS community, we opt for the term "explainability" aligning with Biran and Cotton's

(2017) definition that the explainability of an AI model refers to:

*"The extent to which an observer can comprehend the rationale behind a decision".*

Accordingly, XAI can be characterized as a branch of AI dedicated to elucidating the explanation of a model's decision (Gunning et al., 2019; Miller, 2019; Murdoch et al., 2019; Minh et al., 2022). For clarity in ESS domains, we refine the definition of XAI as:

*"Given an audience, the methods that provide additional insights into an AI model's internal logic during the modeling process beyond the mere outputs".*

This accompanying explanation serves to increase transparency and understanding of the AI model's decisions. Importantly, such explanations should cater to the diverse audiences in the ESS domain and facilitate better access to and understanding of the model's reasoning (Barredo Arrieta et al., 2020).

In this review, we do not delve into the diverse taxonomy of XAI methods (see Appendix A, Dosilovic et al., 2018; Guidotti et al., 2019; Vilone and Longo, 2021; Holzinger et al., 2022), as this is not our primary focus in the context of ESS. The XAI methods in this review are systematically classified and described in Table A1. To align with the practices of the ESS community, we categorize XAI methods into two broad groups: inherently interpretable AI models and methods for explaining black-box models including model-agnostic (MA) and model-specific (MS) methods following Adadi and Berrada (2018) in Figure 2a. MA methods can be applied to a wide range of ML models, while MS methods are tailored and only applicable to specific models.

Inherently interpretable AI models such as linear regression, decision trees, clustering algorithms and variational autoencoder (VAE) are commonly used due to their ability to make people understandable. However, there is a trade-off between accuracy and explainability: inherently interpretable AI models (i.e., linear regression and decision trees) typically perform worse than their less transparent counterparts (Gunning et al., 2021).

Several strategies are typically used to explain black-box AI models, such as surrogate models, perturbation-based methods, and information extraction methods in Figure 2b. These approaches are applicable to both MA and MS methods. The surrogate model involves constructing an inherently interpretable model to mimic the behavior of a complex AI model. Perturbation-based methods explain AI models by perturbing their inputs and observing the subsequent changes in the output (Cortez & Embrechts, 2013). This analysis offers a natural and intuit vim ive way to understand black-box models by focusing on how changes in inputs affect output. Information extraction methods are predominantly model-specific and extract relevant information directly from the model or data, such as visual attention, cell- and layer-level gradients, and latent representations. Although this information is highly valuable, it can be difficult to interpret.

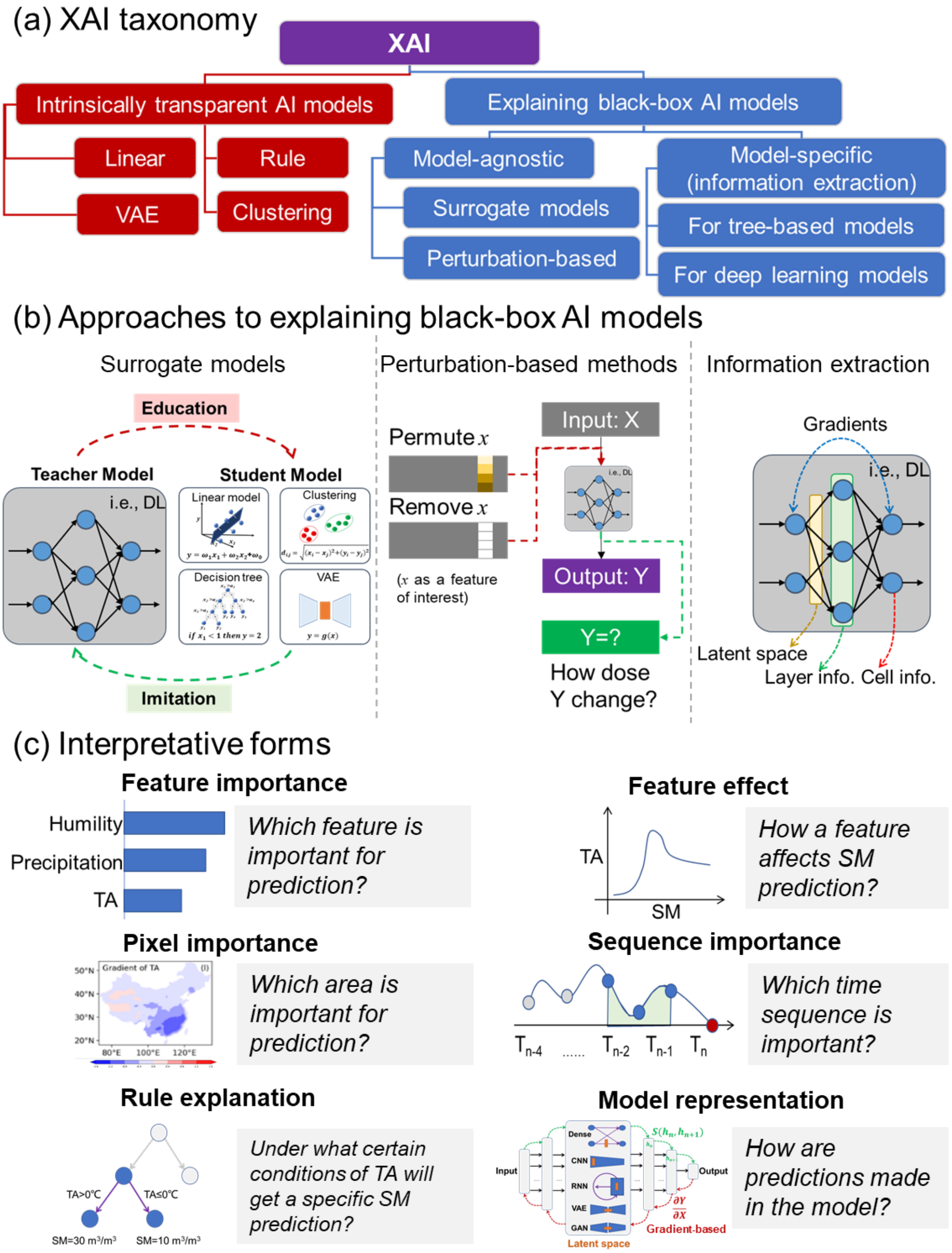

Figure 2. A visual representation of XAI. (a) XAI taxonomy; (b) Approaches for to explaining black-box AI models including surrogate models, perturbation-based methods and information extraction methods; (c) interpretative forms of XAI and the questions they address. This example utilizes uses soil moisture (SM) prediction to demonstrate how different XAI approaches can explain the impact of air temperature (TA) on predictions in distinct ways.

In Figure 2c, we illustrate several common interpretative forms provided by XAI methods. Feature importance is a widely adopted form in ESS that explicitly ranks the relative contribution of features to predictions. Pixel importance highlights the specific image pixels that the AI algorithm focuses on when making a prediction, typically presented as a heatmap to visualize feature importance in space and identify areas where feature variations are critical for accurate predictions. Sequence importance serves the ESS community by identifying critical time periods in historical data that significantly influence future predictions. Feature effect analysis examines how individual features affect predictions; for example, as shown in Figure 2c, soil moisture (SM) exhibits a nonlinear relationship with increasing air temperature (TA). Rule explanation translates the model's decision-making process into human-readable "if-then" statements, clarifying how specific feature values lead to particular predictions. In addition, model representation techniques delve into the internal mechanisms of the AI model and transform them into understandable information that explains how predictions are generated. This information can include representations of forward-propagating processes, back-propagating gradients, or meaningful details extracted from various layers of the neural network.

## 2.2 Explaining Black-box AI Models

### 2.2.1 Model-agnostic Methods

Model-agnostic (MA) methods, which operate independently of the internal mechanisms that generate AI model predictions, can be applied to any type of AI model. MA methods typically use surrogate models and/or perturbation-based methods. Among surrogate methods, local interpretable model-agnostic explanation (LIME) is widely used (Ribeiro et al., 2016). Given an input x (e.g., an image), LIME systematically perturbs the input to generate a set of artificial samples from the neighboring regions. Then, these perturbed instances are fed into a linear regression model, and the feature importance is derived based on the regression coefficients. Based on LIME, several similar methods have been proposed by Guo et al. (2018), Zafar and Khan (2019) and ElShawi et al. (2019).

Among the perturbation-based methods, permutation importance (PI) stands out as a popular choice (Fisher et al., 2021). It provides feature importance across the entire AI model by randomly shuffling the training set data of each variable and measuring the subsequent changes in prediction errors. The perturbation-based technique also allows for the assessment of feature effects, revealing how the model's output responds to variations in input features. This approach is closely related to the concept of sensitivity analysis in ESS. Notable examples include partial dependence plot (PDP, Friedman, 2001), individual conditional expectation (ICE, Goldstein et al., 2015), and accumulated local effect (ALE, Apley & Zhu, 2019). PDP represents the averaged marginal effect of a specific feature on the predictive outcome of the model, while ICE extends this notion to provide local explanations by examining individual instance-specific marginal effects. ALE computes differences in predictions rather than averages over a small window (e.g., using empirical quantiles). As an alternative to PDP, ALE may be subject to bias when dealing with highly correlated variables.

Shapley additive explanations (SHAP), first introduced by Lundberg and Lee (2017), integrates surrogate modeling with perturbation analysis. SHAP uses concepts from coalitional game theory to compute marginal contributions for each feature in a data instance, treating these features as players in a coalition. The marginal contributions are derived by perturbing individual features within an instance. SHAP then adopts additive feature attribution methods as a surrogate model, essentially using linear combinations of input features to construct an interpretable approximation of the original complex model. Following the foundational work on SHAP, several specialized variants have been developed, including kernel SHAP by Covert and Lee (2021), which extends SHAP to kernel-based models; Tree SHAP (Yang, 2022), specifically designed for tree-based algorithms; and DeepSHAP (Chen et al., 2019a), tailored for deep learning architectures.

Other popular MA methods include counterfactuals and prototypes. Counterfactuals illustrate how much a particular instance's features would have to change to substantially influence the prediction (Chou et al., 2022). Prototypes can help to find the important samples in training set data when users want to see what the model considers typical or important (Gurumoorthy et al., 2019).

**2.2.2 Model-specific Methods**

**2.2.2.1 Model-specific Methods for Tree-based AI Models**

Tree-based feature importance (TFI), proposed by Breiman (2001), measures the relevance of variables in tree-based models such as RF, extreme gradient boosting, and gradient-boosting decision tree. TFI assesses feature importance by impurity reduction or disorder that results from randomly perturbating the feature of interest. And the impurity reduction or disorder is aggregated across all trees. A feature is considered important for prediction as it significantly reduces this impurity score when used to split the dataset. TFI has been widely used in ESS modeling to identify and select influential factors. Another interpreter for tree-based models, "Treeinterpreter" tool was developed by Saabas (2015). For any given sample, it extracts the decision path through the forest from the root node to the terminal leaf and calculates the contribution of each predictor, providing a granular explanation of how these models arrive at their predictions.

**2.3.2.2 Model-specific methods for deep learning models**

In the context of model-specific (MS) methods, we focus on XAI techniques tailored for DL models, including XAI methods applicable to all DL models, as well as specific XAI methods for RNN, CNN, and other specialized DL models targeting individual layers or cells.

XAI methods applicable to all DL models include gradient-based methods and the attention mechanism. Gradient-based methods treat DL models holistically and usually use backpropagating gradients through the DL model to estimate feature relevance (more details described in Appendix B). Saliency maps, for instance, compute the gradient of the output with respect to the input image, thereby generating feature attribution maps (Simonyan et al., 2014). This strategy is followed by guided backpropagation (Springenberg et al., 2015), gradient-input (Shrikumar et al., 2019), integrated gradients (Sundararajan et al., 2017), smooth gradient (Smilkov et al., 2017), and expected gradients (Erion et al., 2021). While these methods can produce explicit, relatively robust feature attribution maps, they rely on the differentiability or smoothness of neurons in the model. To overcome this limitation, layer-wise relevance propagation (LRP) is based on the relevance between layers and is propagated from the output to the input, which computationally redistributes the importance across each input dimension of a sample (Bach et al., 2015). Another perturbation-based method, named Occlusion by Zeiler and Fergus (2014), systematically replaces contiguous parts of an input with a baseline value to assess their impact on the prediction function. The Jacobian matrix, similar to gradients for single-output networks, visualizes the backpropagation of errors, showing the differential calculus between inputs and outputs at each layer, thereby indicating the direction of sensitivity (Motteler et al., 1995; Aires et al., 2004; Rahwan et al., 2019).

Additionally, certain methods provide forward propagation representations within DL models. Canonical correlation analysis (Burges, 2010) estimates the similarity between layers to explain DL models. These novel methods based on information similarity have also been used to diagnose DL models (Raghu et al., 2017; Kar et al., 2022). Furthermore, the attention mechanism introduced by Vaswani et al. (2017), is a disruptive innovation that aims to improve the ability of DL models to manage long-range dependencies while enhancing their explainability. This mechanism works by allowing the model to selectively focus its processing on certain segments of the input during task execution. It dynamically weights various input elements, indicating their relative importance or relevance.

Various methods have been developed for RNNs. Some approaches guide RNNs to learn and capture physical relationships (Tang et al., 2022) and spatial patterns (Milisav & Misic, 2023), and enable interpretable sequential data processing (Hou et al., 2021). Other strategies transform RNNs into models that are both globally and locally interpretable by exploiting known theories in terms of time series processing (Wisdom et al., 2016). Probe techniques have also been used to uncover the latent representations within LSTM cells using a linear surrogate model (Lees et al., 2022) or a clustering-based surrogate model (Raghu & Schmidt, 2020). Symbolic aggregate approximation often complements the LSTM by identifying significant points or sequences in time series data, highlighting which time periods are critical for the current prediction (Liu et al., 2023a).

When it comes to CNNs, many XAI methods have been tailored specifically for these architectures. One widely used technique is class activation mapping (CAM), which replaces the fully connected layer with a global average pooling layer in CNNs, thereby identifying important image regions by mapping output layer weights onto convolutional feature maps; channels with higher activations correspond to more informative signals. Gradient-weighted CAM (Grad-CAM), introduced by Selvaraju et al. (2020), refines this concept by generating a weighted class activation map that combines gradients and CAM, offering a coarser-grained visualization. Extensions of CAM, such as U-CAM (Patro et al., 2018), Eigen-CAM (Bany & Yeasin, 2021), and Score-CAM (Ibrahim & Shafiq, 2022), have gained popularity in computer vision applications. Alternative XAI methods modify the loss function to enhance explainability (Springenberg et al., 2015), while others integrate visualization components such as prototype layers (Chen et al., 2019b) or autoencoders (Zhang et al., 2021b; Tavanaei, 2021) to explain CNN functionality.

Explaining some state-of-the-art DL models such as GANs, diffusion models, and Transformers typically involves extracting and interpreting their latent representations. For GANs, dissecting and analyzing the latent space can reveal which units influence the final generated samples (Bau et al., 2019) and help learn more salient features (Chen et al., 2016). The complex latent representations of diffusion models can be extracted through surrogating (Kwon et al., 2022), grouping (Lee et al., 2023), or prototyping (Aghasanli et al., 2023). In the case of Transformers, attention maps (Castangia et al., 2023) offer a means of visualization, although they are only partially interpretable (Clark et al., 2019). To improve interpretability, attention maps have been enhanced with heuristic propagation (Playout et al., 2022) and symbolic abstraction (Schwenke et al., 2023), the latter providing both global and local explanations. As a result, explaining models in terms of both global and local perspectives has become increasingly prevalent.

**2.3 Evaluation of XAI**

To ensure rigorous and verifiable research in XAI, there is a pressing need for comparable evaluation metrics to address the current lack of quantitative evaluation methods. A recent review by Nauta et al. (2023) shows that 58% of technical evaluations in XAI rely on quantitative measures, while 33% use only anecdotal evidence and 22% involve human subject evaluations. The use of quantitative measures facilitates rigorous evaluation by multiple metrics, whereas human evaluation relies on intuitive notions of quality defined by researchers or users. Anecdotal evidence typically depends on what constitutes the “ground truth”, which can typically be derived from synthetic datasets, causal relationships, or domain knowledge.

In the context of ESS, nonlinear functions derived from actual processes (Mamalakis et al., 2022a) and benchmark datasets specifically designed to evaluate explanations (Arras et al., 2022; Mamalakis et al., 2022b) serve as the "ground truth" standards against which XAI methods are measured. Table 2 presents case studies that serve as examples for evaluating XAI methods. The recommended XAI approaches can serve as a preliminary guide for those new to the field.

Table 2. Overview of representative case studies evaluating XAI methods across three evaluation metrics, including quantitativeness (Q), anecdotal evidence (AE), and human subjectivity (HS). For each case, information is provided including research fields, targets, tasks (classification [C] or prediction [P]), input data formats (image [I] or tabular data [T]), and recommended method(s) among those compared.

| Field | Object | Task | Input data format | Recom-mended Method(s) | Metric | Methods in comparison* | Reference |
|---|---|---|---|---|---|---|---|
| Agriculture | Plant disease | C | I | LIME | AE | G-CAM, SHAP | (de Benito Fernández et al., 2023) |
| Remote sensing | Water, forest | C | I | OC, LIME | AE | SA, IG, GI, G-CAM, GG-CAM, GB | (Kakogeorgiou & Karantzalos, 2021) |
| Atmosphere | Atmospheric river | C | I | LRP | AE | SA, SG, PN, IG, GI, DTD, DSHAP | (Mamalakis et al., 2022b) |
| Climate | Temperature | P | I | LRP | AE | SA, SG, DTD, LRP, GI, IG, OC | (Mamalakis et al., 2022a) |
| Climate | Temperature | P | I | LRP | Q | SA, IG, GI, SG, NG, FG | (Bommer et al., 2023) |
| CV | Recognition | C | I | IG | Q | SA, GI, IG, LRP, OC | (Alvarez-Melis and Jaakkola, 2018) |
| CV | Recognition | C | I | IG, DT | Q | DTD, IG, SA, PN, PA, GB, SG | (Kindermans et al., 2019) |
| Medicine | Electroencephalography | P | I | \ | Q | PN, SG, LRP | (Mayor Torres et al., 2023) |
| Medicine | Skin cancer | C | I | IG | Q | IG, SHAP, LIME | (Saarela & Geogieva, 2022) |
| Medicine | Breast cancer | P | T | SHAPG | Q | LR, DT | (Mariotti et al., 2023) |
| Finance | Payment | P | T | SHAP, TR | HS/Q | LIME | (Jesus et al., 2021) |
| Internet | Trojan | P | P | FA | Q | SA, GB, G-CAM, GG-CAM, OC, FA, LIME | (Lin et al., 2021) |
| Health | Insurance | C/P | T | SHAP | Q | PI | (Lozano-Murcia et al., 2023) |
| Sociology | Synthetic data | P | T | SHAP | Q | LIME | (Oblizanov et al., 2023) |
| Internet | Synthetic data | C | T | SHAP | AE | LIME, SA, GI, IG, LRP | (Tritscher et al., 2020) |
| Checkerboard | Synthetic data | C | T | LIME | AE | PI, RSK, SHAP, SK | (Yeo et al., 2022) |

*The abbreviation of XAI methods: LIME (local interpretable model-agnostic explanations), SHAP (Shapley additive explanations), Grad-CAM (gradient-weighted class activation mapping), OC (occlusion), SA (saliency map), GI (input × gradient), IG (integrated gradients), GG-CAM (guided Grad-CAM), GB (guided backpropagation), LRP (layerwise relevance propagation), SG (smooth gradient), PN (PatternNet), DTD (Deep Taylor decomposition), DSHAP (Deep SHAP), NG (noise gradient), FG (fusion gradiant), PA (pattern attribution), SHAPG (ShapGAP), LR (logistic regression), DT (decision trees), TR (treeinterpreter), FA (feature ablation), PI (permutation importance), DL (DeepLIFT), RSK (Robnik-Sikonja and Kononenko), SK (Robnik-Sikonja and Kononenko).

The evaluation of XAI methods based on human subjective judgment and anecdotal evidence in ESS is currently challenging due to the lack of clear ground truth. Therefore, this review focuses more on the quantitative approach. Four critical metrics are considered: faithfulness, stability, robustness, and efficiency, as discussed by Melis and Jaakkola (2018), Murdoch et al. (2019), Minh et al. (2022), and Weber et al. (2023). Faithfulness measures the extent to which an XAI method accurately identifies important features that genuinely influence a model's prediction; if an XAI method assigns high relevance, it should actually change the outcome (Alvarez-Melis & Jaakkola, 2018; Bommer et al., 2023). Robustness is defined as "the persistence of a method for explainability to withstand small perturbations of the input that do not change the prediction of the model" (Alvarez-Melis and Jaakkola, 2018). Huang et al. (2020) emphasize that this metric measures how consistently an explanation remains valid despite small input perturbations. Stability refers to the repeatability of explanations under similar or neighboring inputs; a stable XAI method provides consistent explanations under small input changes (Alvarez-Melis and Jaakkola, 2018). Efficiency concerns the computational cost required to generate an explanation (Adadi & Berrada, 2018; Vilone & Longo, 2021).

In Table 2, SHAP was found to be more advantageous than the others due to its higher faithfulness for tabular data, while IG and LRP more valid when dealing with image data. It is important to note that these evaluations are not definitive, but rather indicative. The suitability of XAI methods may vary depending on the specific use case, so a thorough evaluation is recommended before deployment. To date, there is no one perfect method for all cases. To assist newcomers to the field, several practical XAI packages and evaluation toolkits are provided in Table A2.

## 3. XAI: From Theory to ESS Application

### 3.1 The Needs for XAI

In Section 2.1, we emphasize the importance of considering "for a given audience" in ESS since different stakeholders have varying needs for explanations. The needs significantly influence interpretative form and XAI method selection within specific tasks. In the context of XAI for ESS, audiences can be broadly classified into four primary groups: weather forecasters, policymakers and regulators, AI modelers, and process-based modelers, and geoscientists (in Figure 3). Each group has different needs and goals:

Weather forecasters: these professionals are tasked with assessing the accuracy of forecasts,

essentially answering the question, “Is the forecast accurate and why?”. XAI may provide valuable insight to assist forecasters by comparing predictions with those from numerical weather prediction models or radar extrapolation, along with their expert judgment based on experience. They often require solid evidence to determine what factors contribute to extreme predictions and what indicators signal hazards (Schultz et al., 2021).

Policymakers and regulators: they are responsible for ensuring that AI-enabled policymaking is both reliable and safe and want to know “Is the AI trustworthy?”. They need that the AI model and its explanation are trustworthy. Alternatively, scientific insights from AI can provide essential guidance and evidence for policymaking on issues such as climate change (Dhar, 2020; Haupt et al., 2021). Therefore, regulators need to examine the potential risks beneath the surficial successes of AI. They seek valid, generalizable, and compatible explanations to support informed decision-making.

AI modelers: they focus on uncovering the root causes of poor model performance or generalizability, and need to know “How can we make AI better?”. XAI can reveal biases and errors in AI models that may be due to data quality, model architecture, or training procedures. These insights help developers design, debug, and refine models.

Process-based modelers and geoscientists: they intend to improve our understanding of ESS processes through AI insights focused on the question of “How to get meaningful insights?”. If properly interpreted, AI’s ability to capture non-linearity and coupled states is highly valuable. XAI is potential to help modelers and geoscientists generate hypotheses. For example, the non-linear relationship between evapotranspiration and soil moisture revealed by XAI may help to understand land-atmosphere interactions (Bergen et al., 2019). It emphasizes that the physical insights derived from XAI should be reliable and verifiable. Geoscientists use XAI to explain natural processes, while process-based modelers use it to address limitations in ESMs (Ebert-Uphoff & Hilburn, 2020; Ross et al., 2023). Ultimately, they strive for a fully accurate and robust modeling paradigm for the ESS (Roscher et al., 2020).

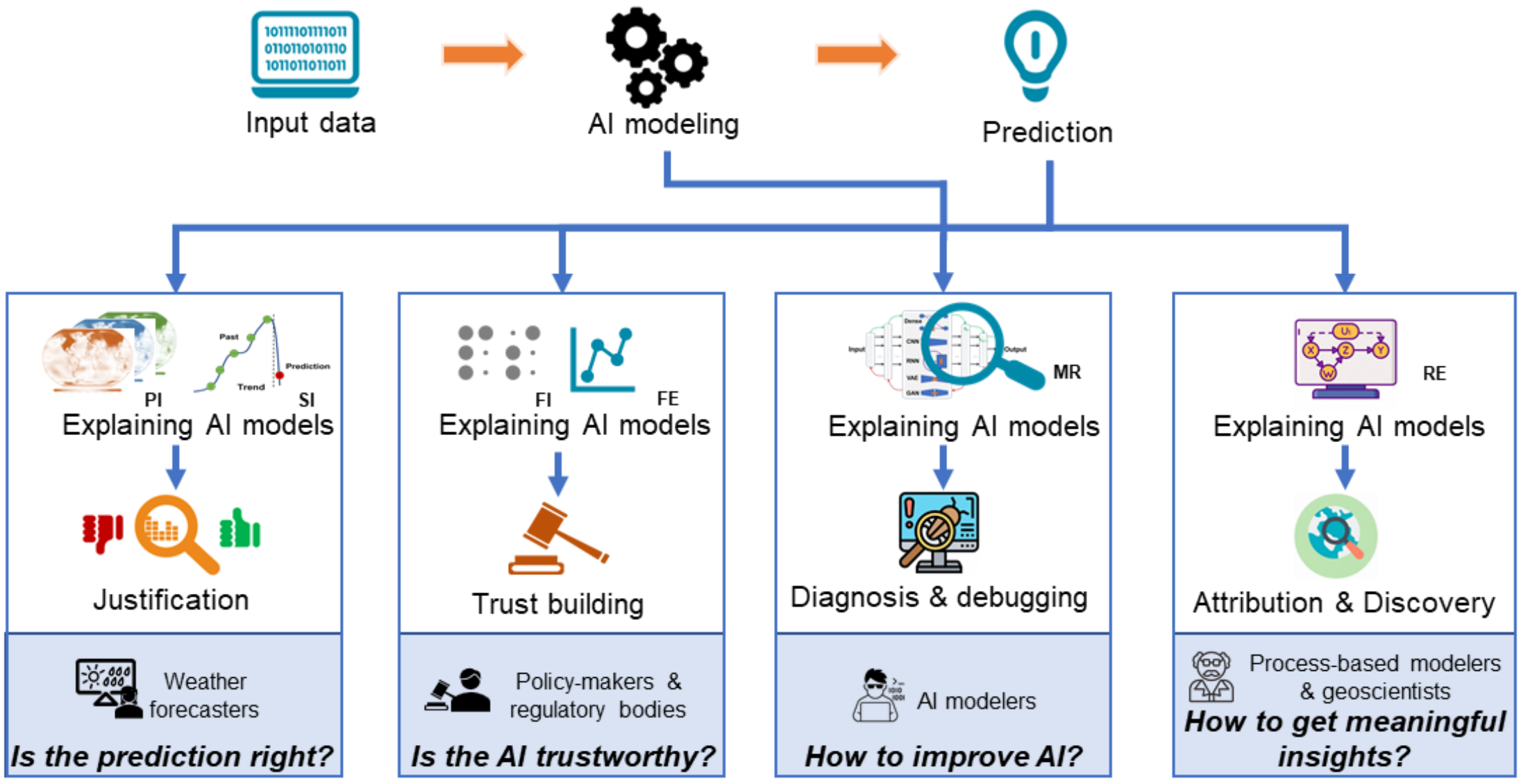

Figure. 3 A schematic overview of the XAI workflow for four audience groups in the context of ESS. The orange lines represent the general machine learning process, while the blue lines highlight the tailored explanation processes for each audience. The blue boxes encapsulate the key questions that these audiences seek to answer using XAI methods. The abbreviations PI, SI, FI, FE, MR, and RE represent pixel importance, sequence importance, feature importance, feature effect, model representation, and rule explanation, respectively—each serving as an example of an interpretative form relevant to a specific

task within XAI for ESS applications.

The four audience groups outlined above each have three central objectives when using XAI: communicating model decisions (for all users), diagnosing and improving models (primarily for AI modelers), and uncovering scientific insights (primarily for process-based modelers and geoscientists). These three aspects are discussed in detail in Sections 3.3 to 3.5.

**3.2 Current States of XAI-ESS**

In this review, we employ a systematic approach to identify and evaluate literature on XAI applications in ESS, as detailed in Appendix C. We retrieved a total of 285 papers, with the use of XAI concepts in ESS research ranging from 2007 to 2023. Figure A2 graphically depicts the annual distribution of these selected papers, showing a marked increase in XAI-ESS publications from 2021 and continuing through 2023.

Figure 4a illustrates the global distribution of study areas and corresponding countries, showing that 44 studies had a worldwide scope. The United States leads with 84 studies, followed by China with 72 studies. Europe, Australia, and India also emerged as significant regions of interest for researchers using XAI methods. A word cloud generated from the titles of the selected papers (Figure 4b) highlights the focus of geoscientists on XAI in ESS. In addition to core topics such as XAI, DL, and ML, "prediction" and "forecasting" were prominent themes, underlining the importance of interpretability in predictive tasks. Terms such as "drivers", "insights" and "factors" further emphasize the need for XAI to provide physical insights within ESS.

The majority of the selected papers pertained to sub-domains including land, hydrology, atmosphere, and hazards. The criteria for dividing these sub-domains are outlined in Figure A3. Researchers predominantly used XAI to address complexities related to water, carbon, and energy flows, exchanges, and cycles between land and atmosphere. Figure 4d indicates that precipitation, streamflow, and temperature have received considerable attention. Moreover, hazards such as droughts, floods, landslides, earthquakes, and wildfires—shown in Figs. 4c and 4d—are complex issues where XAI has potential applications. Agriculture is another important sub-domain within XAI-ESS, with crop yield being the most studied variable, as shown in Figure 4d.

In terms of XAI method types, model-agnostic (MA) methods (57.9%) are the most widely applied due to their high flexibility. Meanwhile, 30.9% and 11.2% of the studies used model-specific (MS) and inherent interpretable AI (IT) methods, respectively. Among them, SHAP was the most popular XAI method in ESS, followed by PI and TFI, all three of which offer feature importance while typically having lower computational cost. Within the MS category, LRP and AM were the two most commonly used techniques.

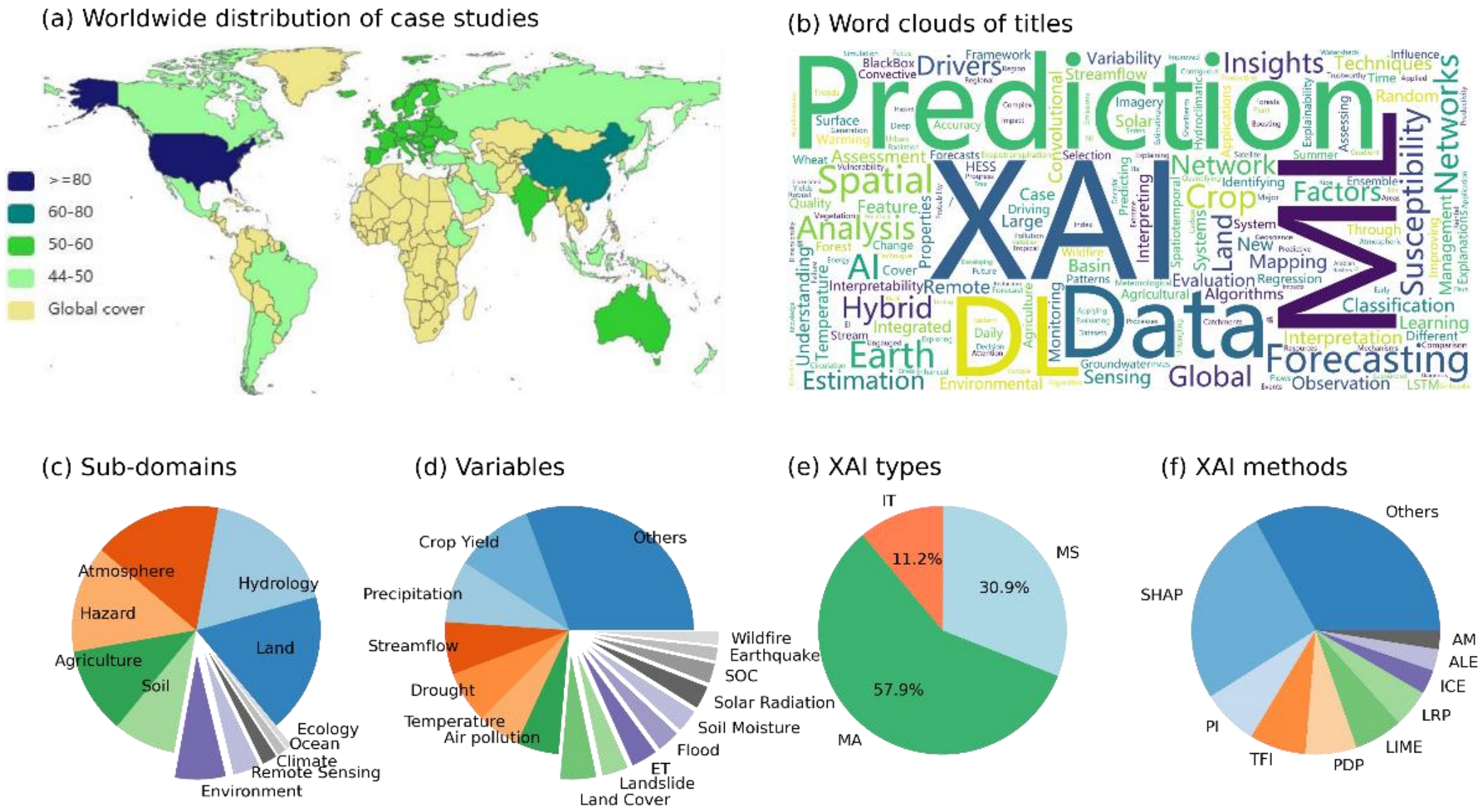

Figure 4. An overview of articles on XAI in ESS from 2007 to 2023. (a) study area, (b) word clouds of acritical titles, (c) sub-domains, (d) targeting variables, (e) XAI types (IT: inherently interpretable AI models; MA: model-agnostic methods; MS: model-specific methods) and (f) XAI methods.

### 3.3 Communicating Model Decisions

XAI plays a critical role in communicating model decisions across various audiences and prediction tasks. Weather forecasters often rely on local feature or pixel importance and rule-based methods to interpret AI-generated forecasts for individual stations or spatio-temporal predictions (Basak et al., 2022; Lim and Zohren, 2021). Feature importance for a specific instance (local) allows them to assess the contribution of specific predictors and validate explanations against their expertise, particularly in time-series predictions for droughts (Feng et al., 2020; Dikshit et al., 2022; Huang et al., 2023a, b), floods (Yang et al., 2020; Ekmekcioğlu et al., 2021), and landslides (Al-Najjar et al., 2022). Pixel importance has also been successfully used to interpret DL models for spatial climate phenomena such as hail (Gagne II et al., 2019), sea temperature, and river discharge (Toms et al., 2020, 2021; Liu et al., 2023b), and oscillation patterns (Schmidt et al., 2020; Gordon et al., 2021; Martin et al., 2022). Rule explanation methods such as surrogate decision trees further aid weather forecasting (Chen et al., 2021).

Policymakers and regulators require high levels of transparency and trustworthiness when using AI-generated predictions, especially in the context of climate change and natural hazards (Haupt et al., 2021; Debnath et al., 2023). Guidelines from UNESCO (2021) and the European Commission's AI Act (2021) require accountability before AI can be used. For instance, DL combined with LRP has been used to provide actionable insights into future climate projections (Diffenbaugh & Barnes, 2023).

For Earth observation researchers using AI, XAI offers a variety of strategies to increase the credibility of AI-generated outputs. Contemporary data products derived from observations require interpretative support to enhance confidence in their use (Shangguan et al., 2017; Li et al., 2022b; Shangguan et al., 2023). Dueben et al. (2022) emphasized that explainability must be integral to the construction of AI-based datasets that are accountable to users. Gevaert (2022) further emphasized the importance of explainability of AI in Earth observation for producers, users, and regulators alike. To date, numerous advanced ESS products have been subjected to interpretation to verify their physical consistency, including remote sensing classification tasks (Stomberg et al., 2021; Hasanpour Zaryabi et al., 2022), as well as products based on observations such as soil moisture (O. & Rene, 2021), soil carbon (Wadoux et al., 2022), surface water levels (Koch et al., 2019), and crop type identification (Orynbaikyzy et al., 2020).

In summary, XAI has demonstrated professional competence in communicating the rationale behind model decisions by extracting explanatory information from the models themselves. This review aims to foster trust in AI models among researchers and policymakers involved in ESS. It advocates the integration of human-in-the-loop AI design, which is not only beneficial for action initiatives such as climate change mitigation but also crucial for effective ESS management.

### 3.4 Diagnosing and Improving Models

Model diagnosis and improvement is a major challenge for AI modelers who seek to understand why an AI model underperforms on specific tasks. Researchers in ESS typically use XAI for feature selection, model determination, and model structure design. Feature selection is essential to ensure the generalizability of a model and to prevent overfitting (Guyon & Elisseeff, 2003; Reunanen, 2003). Unlike traditional filtering methods, which are independent of learning models and rely on data-based evaluation metrics, XAI can directly reveal a model's preference for features, allowing for more informed selection. Feature importance methods such as TFI (Feng et al., 2019; Upadhyaya et al., 2021), PI (Ramirez et al., 2022), and attention mechanisms for DL models (Yan et al., 2021b) are commonly used in ESS research. Studies have shown that XAI-driven feature selection yields better results in prediction tasks (Zacharias et al., 2022; Wang et al., 2023a). Moreover, this approach is consistent with physical processes; Carter et al. (2021) showed how features selected using TFI were consistent with anomalous hydroclimatic circulation patterns when predicting precipitation.

Selecting an appropriate AI model can be both time consuming and computationally intensive. Modelers often assess feature importance in potential models using MA methods to select a model that best fits the underlying physics (Jing et al., 2023). For instance, Schmidt et al. (2020) applied PI to interpret ANN, RF, and linear model in flood forecasting and found varying degrees of agreement between these models and basic hydrological concepts. Hu et al. (2021) compared SHAP explanations of PB, RF, and DL for evapotranspiration and found more similarities between PB and DL suggesting inaccuracies in RF. These findings support a more physically rational model selection.

Identifying biases and errors in model structures can be challenging because adjusting a single component can change the entire model behavior. AI modelers prefer inherently transparent models such as decision trees (Chen et al., 2021) or cubist (Fu et al., 2020). However, they usually result in a loss of model performance and are not suitable for high-dimensional data. Many XAI methods address this issue and enable diagnosis of complex AI models with high-dimensional data. For example, the "treeinterpreter" can pinpoint errors in tree-based precipitation classification models (Upadhyaya et al., 2021), allowing easy identification of problematic tree nodes. In DL models, the Jacobian matrix initially facilitated the removal of less influential connections to profile the atmosphere at high resolution (Blackwell, 2012; Maddy et al., 2021). Shamekh et al. (2023) showed promise in integrating latent representations into AI refining, achieving high accuracy in precipitation intensity prediction.

In summary, XAI is useful for AI modelers optimize performance and build physically consistent models by looking inside black-box models. It has proven effective in aiding data/feature and model selection, thereby enhancing predictive performance in ESS contexts. Ultimately, XAI is demonstrated a valuable asset in the AI model development process.

### 3.5 Providing Scientific Insights

XAI holds great promise for offering physical insights into the ESS due to its remarkable capability of extracting nonlinear relationships. Unlike traditional PB models that generate physical understanding by isolating coupling states or processes, which can introduce substantial systematic bias (Cook et al., 2006; Tuttle and Salvucci, 2016), XAI interprets relationships directly from the data, streamlining the process and conserving computational resources. Here, we showcase some hotspot examples of XAI's success in providing physical insights in ESS sub-domains, including atmospheric science, climatology, oceanography, hydrology and natural hazards.

In atmospheric science, XAI has successfully visualized the complex dynamics of deep convection beneath tropical cyclones (McNeely et al., 2020). Heatmaps generated by XAI have revealed associations between large-scale atmospheric conditions, the occupied area, and the organization degree of deep convection (Retsch et al., 2022). Moreover, the latent representation of a variational autoencoders model has been used to identify different convective regimes in precipitation generation (Behrens et al., 2022).

In climatology, researchers often utilize XAI to pinpoint potential drivers of global warming and extreme precipitation events. They have found that anthropogenic factors contribute to rising temperatures, such as urban green infrastructure (Zumwald et al., 2021) and aerosols/greenhouse gases (Labe & Barnes, 2021), as well as natural factors such as the Pacific decadal oscillation and ocean-atmosphere interactions (Vijverberg & Coumou, 2022). To detect the anthropogenic signal amidst natural variability, Madakumbura et al. (2021), inspired by Barnes et al. (2020), developed an ANN to predict the year associated with annual maps from ESMs. This approach allows XAI to identify patterns of time-varying anthropogenic fingerprints. XAI is also widely used to evaluate the impact of global warming on various aspects of ESS, including carbon dynamics (Patoine et al., 2022), greening trends (Berner et al., 2020), soil respiration (Haaf et al., 2021), soil carbon uptake (Wang et al., 2022c), vegetation (Li et al., 2022c), surface water availability (Webb et al., 2022), groundwater (Chakraborty et al., 2021a; Liu et al., 2022a) and runoff (Anderson et al., 2022).

In oceanography, inherently interpretable linear regression and clustering models are becoming increasingly popular for large-scale modeling of both ocean and atmosphere (Janssens et al., 2021; DelSole & Nedza, 2022). For instance, clustering methods have been instrumental in revealing global ocean dynamical regions, thereby improving the representation of the barotropic vorticity equation, despite the challenges of accurately describing ocean circulation (Sonnewald et al., 2019 and 2023). LRP has also been used to investigate how global heating affects the North Atlantic circulation (Sonnewald & Lguensat, 2021). Sonnewald et al. (2023) noted that insights from XAI could be used to develop a hierarchy of conceptual models of ocean structure and circulation, which could represent an important advance in our understanding of the ocean.

In hydrology, XAI facilitates effective sensitivity analysis and hypothesis generation regarding factors influencing hydrodynamics. For example, it helped identify vegetation (Althoff et al., 2021) and geography (Bai et al., 2022; Liu et al., 2022a) as crucial elements in runoff modeling. SHAP revealed that, when slope is higher than 20%, the runoff is more easily to occur (Wang et al., 2022b). And LIME indicated thresholds for climate variables affecting evapotranspiration (Chakraborty et al., 2021b). Additionally, XAI was used to clarify the parameters of the Budyko framework and to understand hydrological partitioning (Cheng et al., 2022).

Describing and predicting natural hazards including floods, droughts, and wildfires is inherently difficult. XAI helps unravel the underlying mechanisms that may drive these extreme phenomena. In flood analysis, traditional methods often rely on threshold-based techniques (Bertola et al., 2021) or probabilistic frameworks (Blöschl et al., 2019). For example, Stein et al. (2021) used ALE to analyze the nonlinear dynamics of flooding, revealing snowfall and precipitation as significant contributors. Several XAI methods have been used to derive data-driven physical insights into flooding, including SHAP (Yang et al., 2020; Hagen et al., 2021; Aydin and Iban, 2023) and gradient-based approaches tailored for LSTM models (Jiang et al., 2022a, b). In the context of drought studies, XAI has been instrumental in characterizing the generation and properties of droughts (Feng et al., 2019; Saha et al., 2021). For instance, Vidyarthi and Jain (2020) adopted a surrogate decision tree model to elucidate the processes of droughts. In wildfire research, XAI methods are often used to identify which variables significantly influence wildfire behavior. Techniques such as SHAP (Cilli et al., 2022; Abdollahi & Pradhan, 2023; Bountzouklis et al., 2023), IG, and CAM (Monaco et al., 2021; Kondylatos et al., 2022) have been used to discern these sensitivities. For a more comprehensive overview of XAI applications in hazard assessment within ESS, the readers are referred to Ghaffarian et al. (2023).

In summary, XAI has made significant contributions to attribution analysis owning to its ability to fuse information and capture nonlinear relationships (Beucher et al., 2022). This is evidenced by a variety of applications in atmospheric, hydrological, oceanic, and natural hazard contexts within ESS. The ESS community is keenly interested in discovering universal and portable knowledge – phenomena that persist over long periods of time and AI-learned knowledge that is applicable across regions and scales, distilled into fundamental physical understanding. Techniques such as ALE, PDP, and SHAP excel at revealing nonlinear relationships, while LRP and IG highlight spatial correlations among variables. However, rigorous validation of model assumptions and XAI methods is essential to avoid misleading results (Stadtler et al., 2022). The findings of previous studies underscore the importance of considering multiple factors and employing advanced techniques to deepen our understanding of ESS complexities.

## 4. Summary

AI has become an essential component of Earth system science and modelling. The explanation of AI through XAI has been the subject of high expectations in this field. However, there are many details that need to be clarified due to the mismatch between XAI design and ESS background. This review systematically introduces XAI technology to the ESS community, from its theory to ESS applications, while taking into account the ESS background. We present XAI theory, which includes its definition, approaches to explanation, popular techniques, and technical assessment.

We provide a comprehensive literature review on XAI-ESS applications. According to different needs of ESS stakeholders, the current XAI applications can be roughly classified into three functionalities: (1) to communicate with AI decisions through more understandable explanations by demonstrating the physical consistency learned by the AI using XAI methods, (2) to assist AI modelers to diagnose the AI models and guide model building, and (3) to provide scientific insights that helps geoscientists uncover nonlinear relationships hidden in the data and attribution analysis.

However, due to the mismatch between XAI design and ESS requirements, many challenges are associated with the applications. Accordingly, we outline these challenges of using XAI and propose solutions, such as: (1) integrative multiple XAI strategies may help avoid difficulties in selecting the optimal XAI method to ensure faithful explanations; (2) causal physics-informed hybrid XAI modeling could potentially resolve compatibility issues in XAI-ESS arising from varying designs; (3) designing XAI with a human-centered approach for various stakeholders in ESS can address the issues arising from the mismatch between XAI design and the requirements of the ESS community; (4) establishing a quantitative platform for comparison maybe a solution for incommensurability between XAI explanation and physical knowledge.

In the future, we anticipate that XAI to flourish in the field of ESS. Integrated XAI techniques that are more faithful and lighter will be popular in this sub-domain. We also expect more effective XAI methods to be introduced in ESS, especially for diagnosing AI models. Ultimately, we recommend that XAI can effectively help with many issues in ESS, such as new knowledge, data, hybrid modeling, and difference in ESMs, human-computer interaction, and policy proposals.